\documentclass[10pt,a4paper]{article}
\usepackage[left=25mm,right=25mm,top=22mm,bottom=22mm]{geometry}
\usepackage{algorithmic, algorithm}
\usepackage{amssymb}
\usepackage{url}
\usepackage{subfig}
\usepackage{subfloat}
\usepackage{graphicx}
\newtheorem{def1}{Definition}
\begin{document}

\title{Interactive Reference Point-Based Guided Local Search for the Bi-objective Inventory Routing Problem}
\author{Sandra Huber $^1$ \and Martin Josef Geiger $^1$ \and Marc Sevaux $^2$\\[3mm]
 \and $^1$ Logistics Management Department \\
 Holstenhofweg 85, 22043 Hamburg, Germany\\
   {\{sandra-huber,m.j.geiger\}@hsu-hh.de}\\[3mm]
  \and $^2$ Universit\'{e} de Bretagne-Sud\\
  Lab-STICC -- Centre de Recherche\\
  2 rue de St Maud\'{e}, F-56321 Lorient, France\\
  {marc.sevaux@univ-ubs.fr}\\}
\date{}


\maketitle

\begin{abstract}
Eliciting preferences of a decision maker is a key factor to successfully combine search and decision making in an interactive method. Therefore, the progressively integration and simulation of the decision maker is a main concern in an application. We contribute in this direction by proposing an interactive method based on a reference point-based guided local search to the bi-objective Inventory Routing Problem. A local search metaheuristic, working on the delivery intervals, and the Clarke \& Wright savings heuristic is employed for the subsequently obtained Vehicle Routing Problem. To elicit preferences, the decision maker selects a reference point to guide the search in interesting subregions. Additionally, the reference point is used as a reservation point to discard solutions outside the cone, introduced as a convergence criterion. Computational results of the reference point-based guided local search are reported and analyzed on benchmark data in order to show the applicability of the approach.
\end{abstract}

\section{Introduction}
Distribution logistics plays an important role in companies. In this context, the classical Vehicle Routing Problem minimizes distribution costs in order to service the placed customer orders. The class of Inventory Routing Problems (IRP), an important extension of the Vehicle Routing Problem, simultaneously takes into consideration the supply chain management aspects `inventory management \emph{and} transport processes'. In other words, the supplier decides about the delivery quantities and dates of each customer and the subsequent routing. Benefits of this strategy are that the supplier is able to improve the routing while the customer delegates the responsibility of the inventory management to the supplier. The increasing importance of the IRP is e.\,g.\ illustrated in~\cite{coelho:2012}.


While the majority of the literature on the IRP employs single-criteria models~\cite{solyal:2012}, we consider a bi-objective IRP with two separated objective functions, minimizing the `total sum of inventory levels at the customers' over all periods' \emph{and} the `sum of all distances traveled by the vehicles over all  periods'. This formulation seems suitable to deliver the trade-offs between these two aspects to the decision maker (DM)~\cite{geiger:2011}.

In order to find a most-preferred solution to the bi-objective IRP, external information of a decision maker is required. For successfully involving the DM, we assume that (s)he possesses reliable expert knowledge about the investigated problem, and that (s)he is capable of giving preference information in outcome space. On that basis, the chosen solution depends on the preference information of the DM~\cite{eskelinen:2010}.

The paper is structured as follows: After illustrating aspects of decision making in Section~\ref{sec:aspects}, a brief description of the investigated IRP is given in Section~\ref{sec:problem}. The methodology is presented in Section~\ref{sec:method}. Section~\ref{sec:experiments} provides the experimental setup and compares the results of the different reference points with a non-interactive, offline approach. Finally, we present conclusions and some remarks for future work in Section~\ref{sec:future}.

\section{\label{sec:aspects}Aspects of decision making}
\subsection{Combining search and decision making}


Basically, three methods are illustrated in the literature to solve multi-objective problems that combine optimization and decision making: \emph{a priori}, \emph{a posteriori} and \emph{interactive} articulation of preference information~\cite{eppe:2011a,luque:2012}.

Generally \emph{a priori} methods elicit preference information prior to search. Partial utility functions are formulated for the different objectives and combined into a global utility function. If typical rationality axioms hold, and the utility function of the decision maker can be accurately formulated, then e.\,g.\ multiattribute utility theory (MAUT) provides an applicable way of solving multi-objective problems~\cite{roy:1990}. With regard to global utility functions, `compensation' issues must be taken into consideration. Also, the set of objective functions must be exhaustive~\cite{vincke:1990}. In cases of uncertain preferences, such an approach can become problematic. Strictly speaking, this stands against \emph{a priori} methods for modeling preferences in many practical situations.

Contrary to \emph{a priori} methods, the DM is not needed during the search procedure of an \emph{a posteriori} method. Moreover, the decision maker gives his/her preferences after search~\cite{eppe:2011a}. It is suggested that the DM might have difficulties to handle the large amount of data, and the question arises how to support the decision maker in finding the most-preferred solution $x^*$ of the Pareto-set $P$ ($x^* \in P$). Besides, the computational effort might be expensive for generating the set of all Pareto-optimal solutions. It seems also questionable, if \emph{a posteriori} approaches can satisfy changing requirements of the decision maker. The main advantage is that the search and the decision making can be treated separately~\cite{geiger:2008}. According to the aforementioned drawbacks of \emph{a priori} and \emph{a posteriori} methods, our emphasis in this study lies on \emph{interactive approaches} (see Section~\ref{sec:interactive} below).

\subsection{Constitutive elements of multi-criteria decision making}


For the later considered bi-objective IRP, the following well-known definitions apply (minimization of the objective functions $g_{k}$):

\begin{def1}[{\label{def:dominance}}Dominance]
An alternative $x_i \in \mathcal{X}$ dominates an alternative $x_j \in \mathcal{X}$, iff $g_k{(x_i)}$ $\leq$ $g_k{(x_j)}$ $\forall k = 1, \ldots, K$ and $\exists k |\ g_k{(x_i)} < g_k{(x_j)}$. We denote the dominance of $x_i$ over $x_j$ as: $x_i \vartriangleright x_j$.
\end{def1}
\begin{def1}[{\label{def:paretooptimality}}Pareto-optimality]
An alternative $x_i \in$ $\mathcal{X}$ is said to be Pareto-optimal, iff $\nexists x_j \in$ $\mathcal{X} |\ x_j \vartriangleright x_i$.
\end{def1}
\begin{def1}[{\label{def:weakly}}Weakly-efficiency]
An alternative $x_j \in \mathcal{X}$ is weakly-efficient (weakly Pareto-optimal) if there is no other alternative $x_i \in \mathcal{X}$ such that $g_k{(x_i)} < g_k{(x_j)}$ for all $k=1,\ldots,K$~\cite{ehrgott:2005}.
\end{def1}
For the description of Expression~\ref{eqn:cheb} in Section~\ref{sec:method}, we use the concept of weakly-efficient solutions. Following relation between Pareto-optimality and weakly-efficiency exists: the set of Pareto-optimal alternatives are a subset of the set of weakly-efficient alternatives~\cite{miettinen:1999}.

The majority of the literature assumes for preference modeling the capability of the decision maker to state basically three preference relations~\cite{ozturk:2003,vincke:1990}.
\begin{itemize}
\item Preference ($\prec$) of alternative $x$ over $y$, denoted as $x \prec y $,
\item indifference ($\sim$) between $x$ and $y$, denoted as $x \sim y $, and
\item incomparability ($J$), denoted as $x J y \Leftrightarrow \neg (x \prec y) \wedge \neg (y\prec x) \wedge \neg (x \sim y)$.
\end{itemize}

As a remark, the later described methodology in Section~\ref{sec:method} only investigates non-dominated alternatives. Consequently, the DM only evaluates non-dominated solutions in terms of preference relations. Incomparability, as a third preference relation, can be present in case of rather different alternatives. In such situations, the DM might not be able to characterize the alternatives by $\prec$ or $\sim$. \cite{ozturk:2003} gives an overview about how to handle incomparability. The authors also comment on additional issues, such as strength of preferences and others.

\subsection{\label{sec:interactive}Interactive integration of the decision maker}
An interactive method might overcome the disadvantages of a priori and a posteriori methods by gradually specifying preferences of the decision maker. This leads to an interaction of the search and decision making process~\cite{vanderpooten:1989} with the goal of computing and presenting only a preferred subset of solutions to the decision maker.

In this context, the decision maker has the advantage to learn about the feasibility of the solutions of the problem and gain realistic expectations. Trade-offs among the outcomes of the solutions can be shown, and extreme solutions can be presented to the DM~\cite{klamroth:2008}.

Naturally, these `basic' preference parameters are applied: 1) \emph{weighting vectors} demonstrating the relative importance of each criterion, 2) \emph{reservation levels} representing the minimal requirements on each criterion and/or 3) \emph{reference points} expressing aspiration levels or, in other words, the desirable levels of each criterion~\cite{vanderpooten:1989}. Several preference parameters can be used simultaneously in interactive methods. In typical applications, the DM will be only able to state one type of preferences. The aim of the interactive method is guiding the search of the decision maker rather than understanding the precise structure of the preferences~\cite{vanderpooten:1989}. Consequently, an interactive method helps to develop preferences and should be able to deal with inconsistent statements.

It is also common to distinguish between the explicitly and implicitly determination of preference parameters in a model~\cite{vanderpooten:1989}. For example the DM only gives holistic preferences for known situations and then the parameters of the model are estimated implicitly~\cite{eppe:2011}. Contrary, if a DM states a reference point, this aspect is integrated in the model explicitly.
\section{\label{sec:problem}Problem description of the investigated IRP}
For an overview of the IRP, we may also refer to~\cite{coelho:2012}. We observe a single-item IRP in a distribution network with one depot, from which a set of $n$ customers are delivered. At the depot, inventory costs and capacities are not taken into account (they are, however, at each customer $i$). A homogeneous fleet of capacitated vehicles is used and the fleet size is unconstrained.

The delivery strategy is that the customers are served if their current inventory is about to become insufficient. This implies that the demand at each customer $i$ is either fully covered by currently held inventory, or the inventory is zero. This concept ensures that the exact currently demand is replenished. For example: when the inventory level of customer $i$ is $2$ at period $t, t=t-1$ and the demand of customer $i$ at period $t$ is $8$ (note that the inventory level and the truck capacity are not exceeded), the replenishment strategy is to deliver the exact demand of $8$. The idea behind this replenishment strategy is that the customer might not be delivered in the next period resulting in lower routing costs. This situation is solved over a finite planning horizon $T$ with deterministic consumption rates for each customer $i, i = 1,\ldots,n$ and each period $t,  t =1,\ldots,T$.

Decision variables of the investigated IRP are: (i) the delivery quantities $q_{it}$ for each customer $i, i = 1,\ldots,n$ and each period $t$ of the planning horizon $T$ and (ii) the VRP must be solved for each period $t,  t =1,\ldots,T$, including the delivery quantities $q_{it}$ into tours for the involved vehicles. This circumstance explains the overall complexity of the problem at hand.

In our multi-objective formulation of the IRP, two objectives, which are clearly in conflict to each other: (i) inventory levels (sum of all inventory levels at the customers at the end of each period), and (ii) routing costs (sum of all distances traveled by the trucks in each period) are to be minimized simultaneously. According to the trade-off: While small delivery quantities lead to low inventory levels over time, large delivery quantities allow a minimization of the routing costs. For a detailed formulation we refer to~\cite{geiger:2011}.

\section{\label{sec:method}Proposed methodology}
In general, our reference point-based guided local search and the offline approach are based on the ideas of Geiger and Sevaux~\cite{geiger:2011} where the decisions of the IRP are separated into two levels to develop a better understanding of the approach:
\begin{enumerate}
  \item Determination of the delivery quantities for each period.
  \item Computation of the routing for each period, taking the previously computed delivery quantities as an input. Due to the complexity of this underlying Vehicle Routing Problem, a classical savings heuristic~\cite{clarke:1964:article} is used here.
\end{enumerate}

\begin{figure}
\centering
 \subfloat[Explanation of $R_j$.\label{fig:explanation}]{\includegraphics[width=6cm]{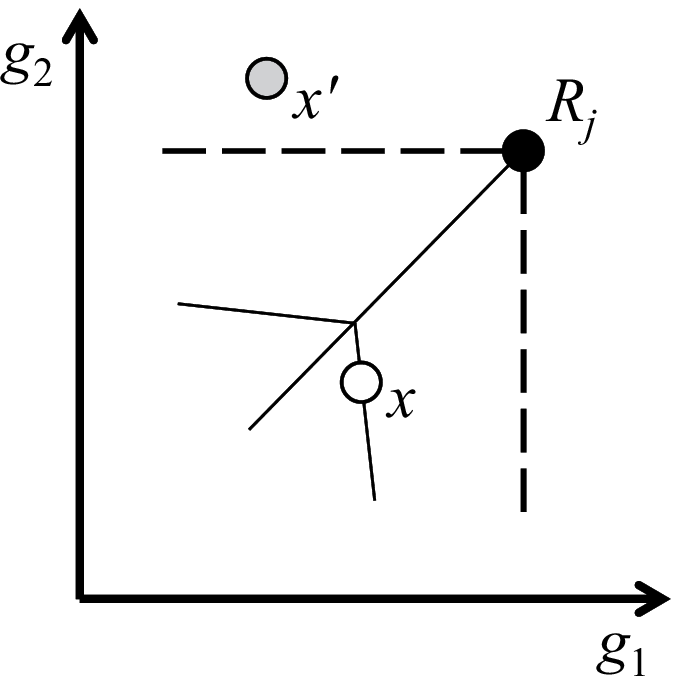}}
\centering
\subfloat[Local search leaving the cone defined by $R_j$.\label{fig:integrationreservationpoint}]{\includegraphics[width=6cm]{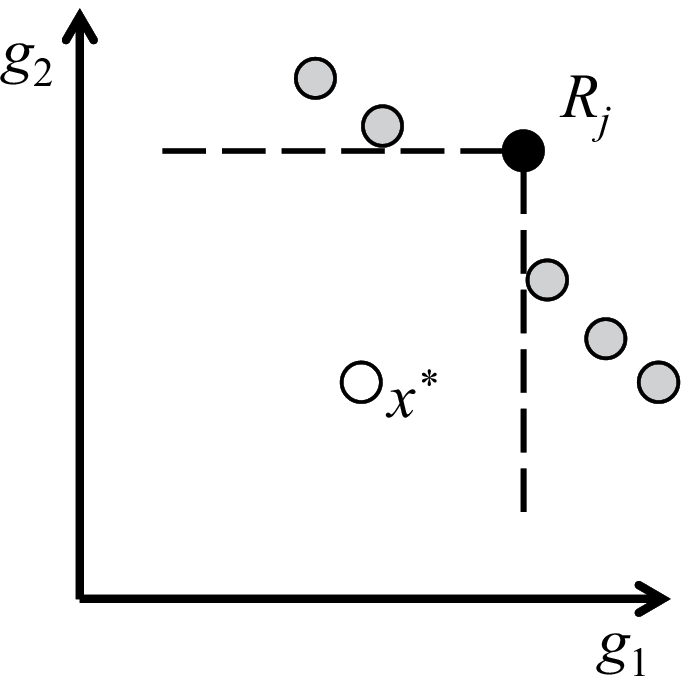}}
 \caption{Using $R_j$ as a reference point and a reservation point~\cite{huber:2013}.}\label{fig:reservationpoint}
\end{figure}

{\bf{Construction procedure:}} describing the reference point-based guided local search (`RPGLS') and the offline approach, we implement an $n$-dimensional vector $\pi = \left( \pi_1, \ldots, \pi_i,\ldots ,\pi_n \right)$. This $n$-dimensional vector represents for how many periods the demand of the customer $i$ is covered (`delivery period'). Only integer number of period demands are delivered, and a fixed number of coverage periods $\pi_i$ is chosen for each customer. E.\,g.\  when $\pi_i = 2$, the exact demand of the customer $i$ is served for the next two consecutive periods. The initial solution assumes a replenishment strategy with identical delivery periods. Starting with $1$ and increasing them by steps of $1$ until the alternative cannot be added to the archive of non-dominated solutions.

{\bf{Improvement procedure:}} a local search run is performed on $\pi$. In the simplest form, a multi-point hillclimbing algorithm can be used, changing the values within $\pi$ by $\pm 1$. Obviously, values $<1$ are to be avoided. Throughout search, an unbounded archive on (currently) non-dominated solutions is kept (preliminary experiments show that the memory of typical current computer hardware is sufficient to store the identified alternatives in the investigated problem). The archive helps deleting solutions by dominance comparisons~\cite{geiger:2011}. In this sense, the method keeps track of all non-dominated solutions. The resulting capacitated Vehicle Routing Problem is then solved with a classical savings heuristic~\cite{clarke:1964:article}.

To direct the search, each selected point in outcome space is then taken as a reference point $R_{j}=(r_{j1},\ldots r_{jk})$, and the goal of the local search algorithm is to minimize the maximum distance of the computed solutions $x$ to this point (see Figure~\ref{fig:explanation}).

Expression~(\ref{eqn:cheb}) defines the distance of each solution $x$ to the reference solutions $R_{j}$. A normalization of the objective function values is performed by means of weights $w_{k}$. For further description of $w_{k}$ see Section~\ref{sec:results}.

\begin{equation}\label{eqn:cheb}
\min \max_{k=1,\ldots,K} \left[ \left\{ w_{k} \left( g_{k} (x) - r_{jk} \right) \right\} + \epsilon \sum_{k=1}^K g_{k}(x)\right]
\end{equation}

Note that, in the general case, the additional term $\epsilon \sum_{k} g_{k}(x)$ is needed to avoid the identification of weakly-efficient solutions. If however the local search algorithm maintains an archive on non-dominated solutions throughout search, and eliminates weakly-efficient solutions by means of dominance comparison (which is what we do in our experiments), then this term can be neglected. \emph{Epsilon} could be determined for this application, but maintaining an archive of non-dominated solutions comes with another effect: despite primarily minimizing expression~(\ref{eqn:cheb}), search can even continue after identifying a (locally) optimal alternative to~(\ref{eqn:cheb}). Then however, search is expected to diverge from the defined direction, spreading out on the Pareto-front.

The main difference between the proposed offline approach and the investigated interactive approach is that the offline approach does not discriminate between different search directions. Instead all computed reference points (initial solution with identical delivery periods) are taken in order to approximate the Pareto-front. Contrary, the DM can state preference information in our interactive approach. The additional contribution, compared with the offline approach, is to guide the search in interesting subregions with the goal of speeding up the solution process. This might come with the effect of a better solution quality.

After guiding the search by selecting a reference point $R_j$, at some later stage, the local search algorithm spreads out of the cone defined by the reference point (see Figure~\ref{fig:explanation}). In an interactive setting, this point in time can be seen as the reaching of a `natural' termination criterion~\cite{huber:2013}.

Figure~\ref{fig:integrationreservationpoint} demonstrates the situation when the local search algorithm `jumps' out of the cone at the beginning of the local search: starting from the reference point $R_j$, the grey alternatives are computed as neighboring solutions. If the search now concentrates on elements within the cone defined by $R_j$, this could imply that the algorithm prematurely terminates, and that no or only a few solutions are found. Therefore, stopping the search when leaving the cone defined by $R_j$ should only be done at a later stage of the local search runs.
\section{\label{sec:experiments}Experimental investigation}
\subsection{Benchmark data and eliciting of preferences}
The proposed methodology has been tested on two benchmark instances proposed by Sevaux and Geiger (2011)~\cite{geiger:2011b}, available under \url{http:\\logistik.hsu-hh.de\IRP}. For both instances, the time horizon is $T = 30$ and the number of customers varies from $50$ ({\tt{GS-01}}) to $75$ ({\tt{GS-02}}). Scenario `a' is chosen in the datasets.  Here, the average demand is constant over time. However, the actual demand in each period can vary $\pm 25\%$ around the average~\cite{geiger:2011}.

We simulated different types of decision makers for the interactive approach. After presenting the DM a first, rough approximation, (s)he guides the search by choosing one preferred search direction. For example, a DM being in favor of low inventory levels and therefore more frequent deliveries would take reference point $R_{1}$, and contrary, a DM preferring higher inventory levels and less frequent deliveries would select $R_{7}$/$R_{8}$ (see Table~\ref{tbl:referencepoints} for an illustration of the reference points vectors). The selected reference point is then used to guide the search into a subregion of the Pareto-front. For each instance, seven respectively eight reference points were tested as illustrated in Table~\ref{tbl:referencepoints}. Note that the number of reference points can differ for the different test instance.

\begin{table}[!ht]
\caption{Reference point vectors of {\tt{GS-01}} and {\tt{GS-02}}.}
\label{tbl:referencepoints}
\smallskip
\centering
\begin{tabular}{|c|c|l||c|c|l|}  \hline
 instance & RP & vectors & instance & RP & vectors\\\hline\hline
{\tt{GS-01}}& $R_1$&  ($r_1 = 2401, r_2 = 12734.9$)& {\tt{GS-02}}& $R_1$ &($r_1 = 4242, r_2 = 17453.3$)\\\hline
{\tt{GS-01}}& $R_2$ &  ($r_1 = 12838, r_2 = 8868.3$)& {\tt{GS-02}}& $R_2$ &($r_1 = 22538, r_2 = 11777.8$) \\\hline
{\tt{GS-01}}& $R_3$ &  ($r_1 = 23377, r_2 = 6669.5$) & {\tt{GS-02}}& $R_3$ &($r_1 = 41101, r_2 = 8965.3$)\\\hline
{\tt{GS-01}}& $R_4$ &  ($r_1 = 33842, r_2 = 5346.3$) & {\tt{GS-02}}& $R_4$ &($r_1 = 59212, r_2 = 7316.3$) \\ \hline
{\tt{GS-01}}& $R_5$ &  ($r_1 = 42390, r_2 = 4696.9$) &{\tt{GS-02}}& $R_5$ &($r_1 = 74637, r_2 = 6852.7$) \\ \hline
{\tt{GS-01}}& $R_6$ &  ($r_1 = 52234, r_2 = 4291.9$) & {\tt{GS-02}}& $R_6$ &($r_1 = 91912, r_2 = 6272.0$) \\ \hline
{\tt{GS-01}}& $R_7$ &  ($r_1 = 64968, r_2 = 3850.3$) & {\tt{GS-02}}& $R_7$ &($r_1 = 113818, r_2 = 5765.4$) \\ \hline
& & & {\tt{GS-02}}& $R_8$ &  ($r_1 = 124820, r_2 = 5659.9$) \\ \hline
\end{tabular}
\end{table}

\subsection{\label{sec:results}Results and discussion}
When analyzing the results of {\tt{GS-01}} and {\tt{GS-02}}, both objective functions are normalized because normalized values are better in terms of quality comparisons of different test instances. This procedure is performed to align the outcomes of the two criteria, which, for themselves, are measured on rather different scales (inventory levels versus traveled distances). This brings us back to expression~(\ref{eqn:cheb}), where the values of $w_k$ are employed to compare the relative importance of the different objectives. It was our intuition to use the maximum and minimum values of the computed outcomes for determining the $w_j$-values, thus normalizing the two criteria.

Once knowing the best-known outcomes from the offline approach, a most-preferred solution can be computed with respect to each reference point $R_{j}$. On the basis of this solution, the distance between the currently found alternatives of the RPGLS and this alternative can then be computed (called `min max weighted Chebyshev'). This distance assumes a value of $0$ (once the most-preferred alternative is found). In some cases, this distance can even assume a negative value. This happens when the most-preferred solution (given by the offline approach) is surpassed by the interactive approach.
\begin{figure}[!ht]
\centering
\subfloat[{\tt{GS-01}}]{\includegraphics[width=8.0cm]{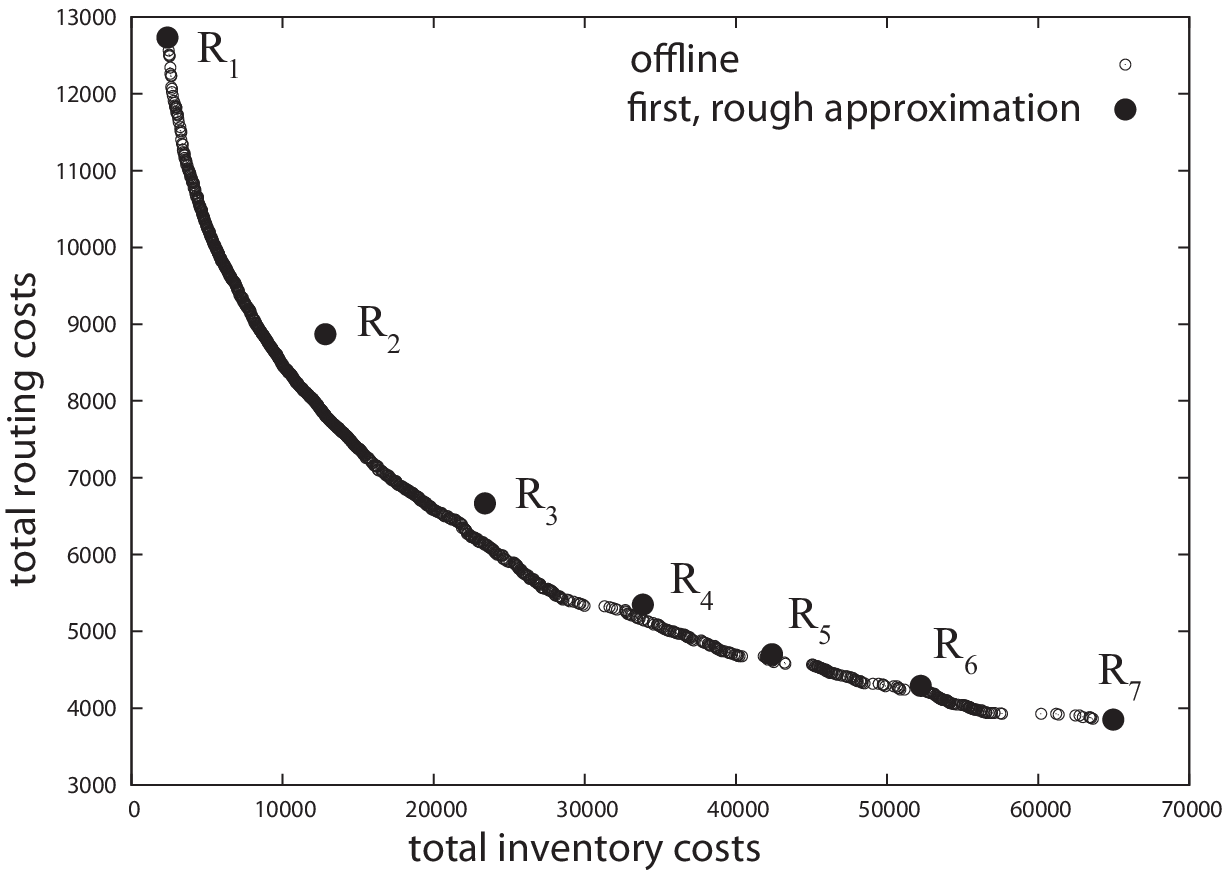}}
\subfloat[{\tt{GS-02}}]{ \includegraphics[width=8.0cm]{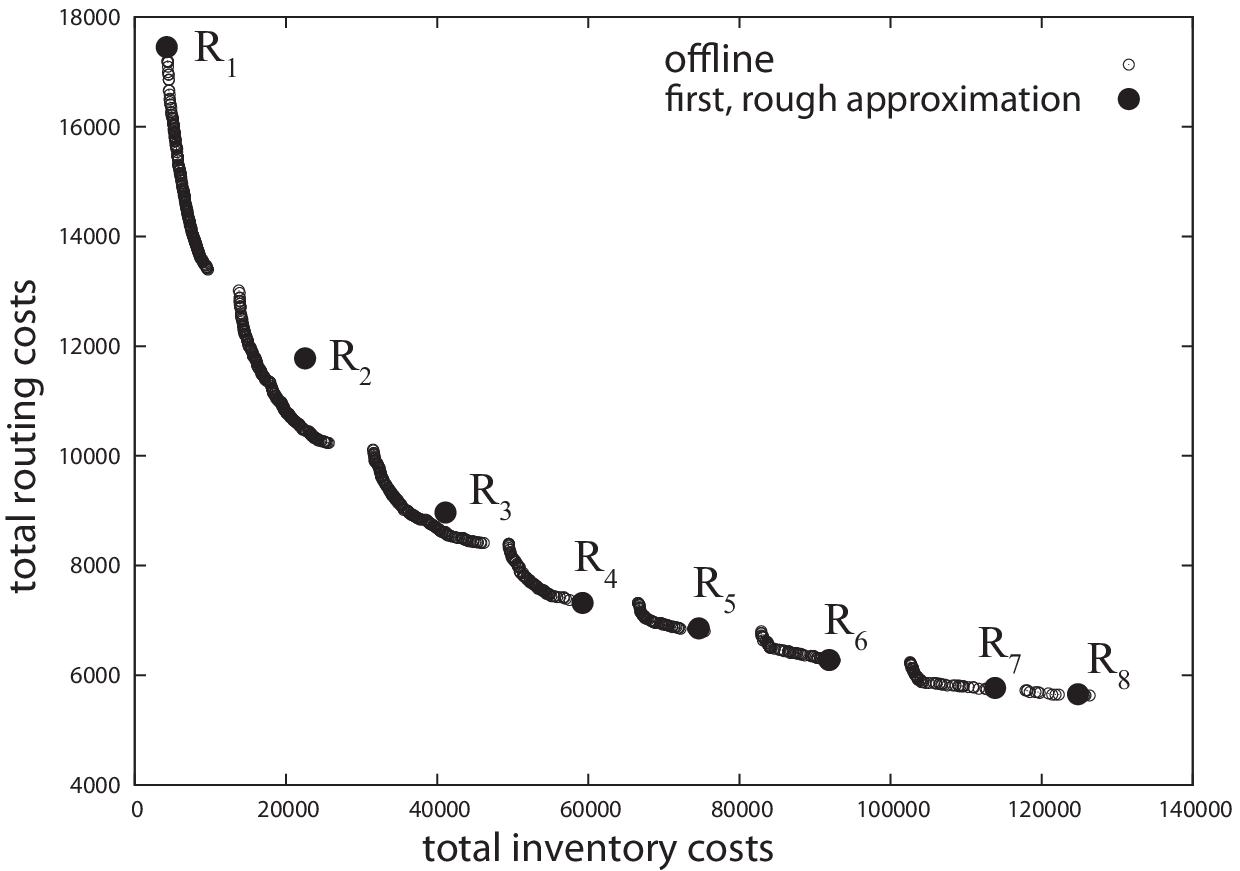}}
 \caption{Results of the first approximation and the offline approach of the Pareto-front.}\label{fig:start}
\end{figure}

The bold black dots in Figure~\ref{fig:start} illustrate the first, rough approximation of the Pareto-front which represents the replenishment strategy of identical delivery periods. Note that this approximation is used for the interactive and the offline approach. Positive is, that the approximation of $P$ is identified fast which is helpful for the interaction of the DM with the search procedure. For example, the computational time for the approximation is 0.146 seconds for {\tt{GS-01}} and 0.457 seconds for {\tt{GS-02}}.

Note that every experiment has been conducted on a single core of an Intel Xeon X5550 processor.

Also the results of the offline approach are described in Figure~\ref{fig:start} (small black circles). This approach takes the initial solution, seven respectively eight reference points, at once and tries to converge to the entire Pareto-front without discriminating between particular search directions. Particularly, they are simultaneously used as an input to approximate the Pareto-front. After representing the DM the results of the offline approach (small black circles), (s)he can select a most-preferred solution. However, the DM might be overwhelmed by the the number of solutions. The outcomes of this approach are plotted for {\tt GS-01} after 838{,}000 evaluations and {\tt GS-02} after $10^6$ evaluations. The computational time for the offline approach is approximately $5$ hours for {\tt GS-01} and approximately $17$ hours for {\tt GS-02}.

\begin{figure}
\begin{minipage}[hbt]{8cm}
  \centering
  \includegraphics[width=8cm]{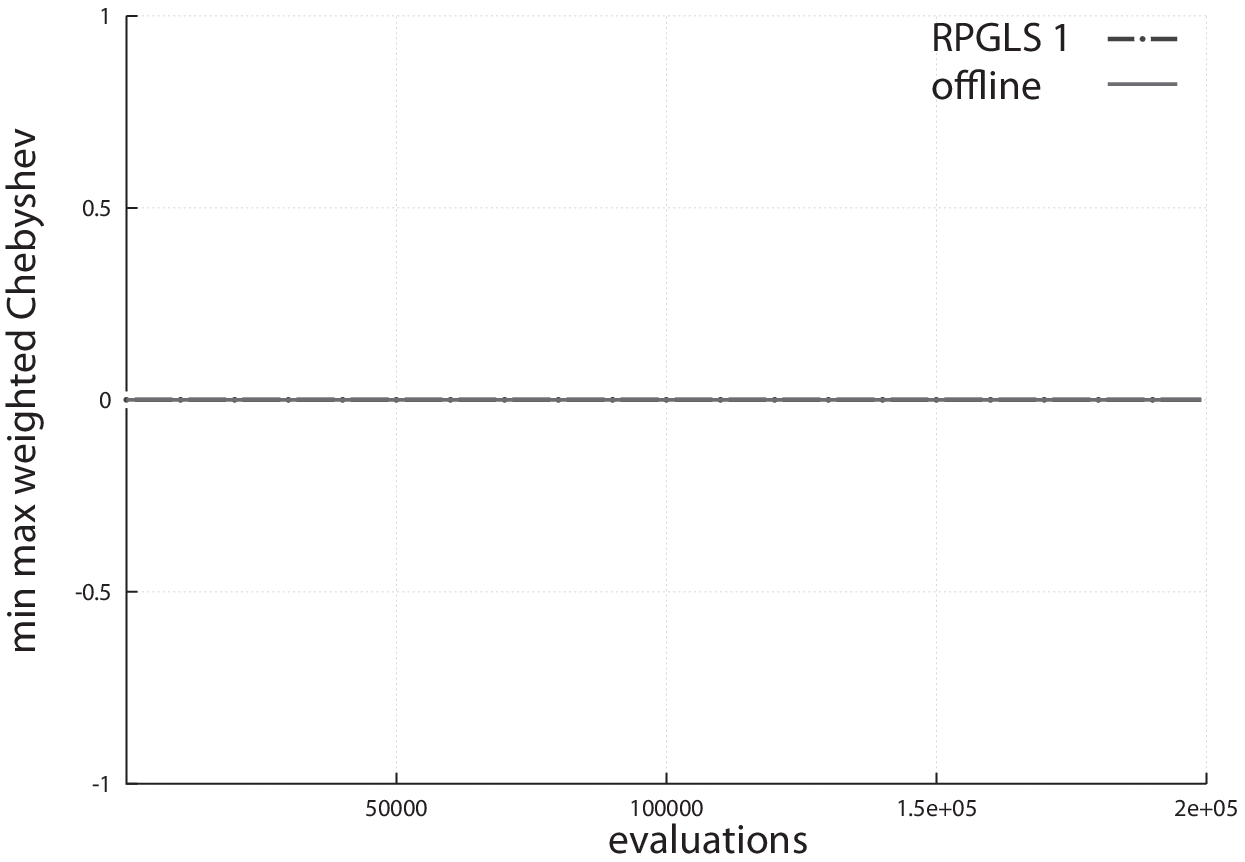}
\end{minipage}
\begin{minipage}[hbt]{8cm}
\centering
  \includegraphics[width=8cm]{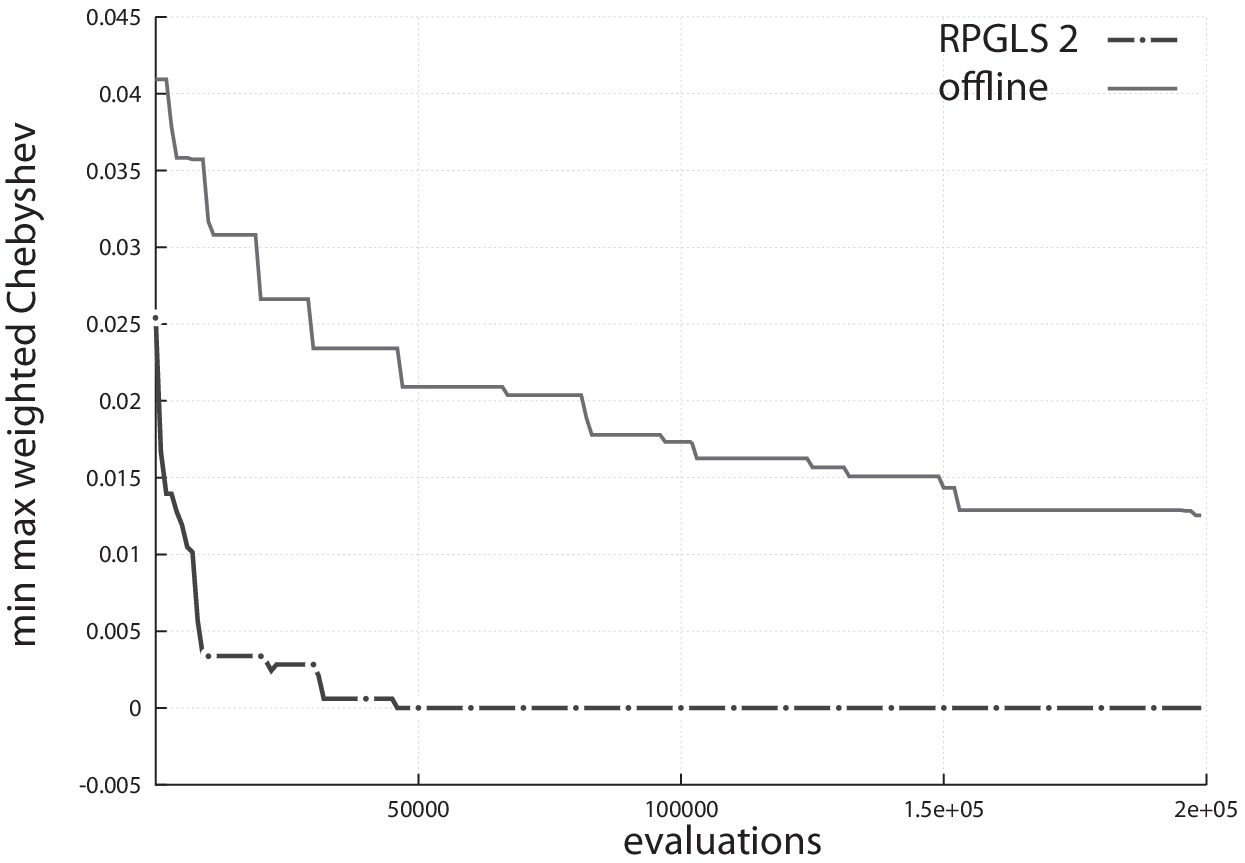}
\end{minipage}

\begin{minipage}[hbt]{8cm}
  \centering
  \includegraphics[width=8cm]{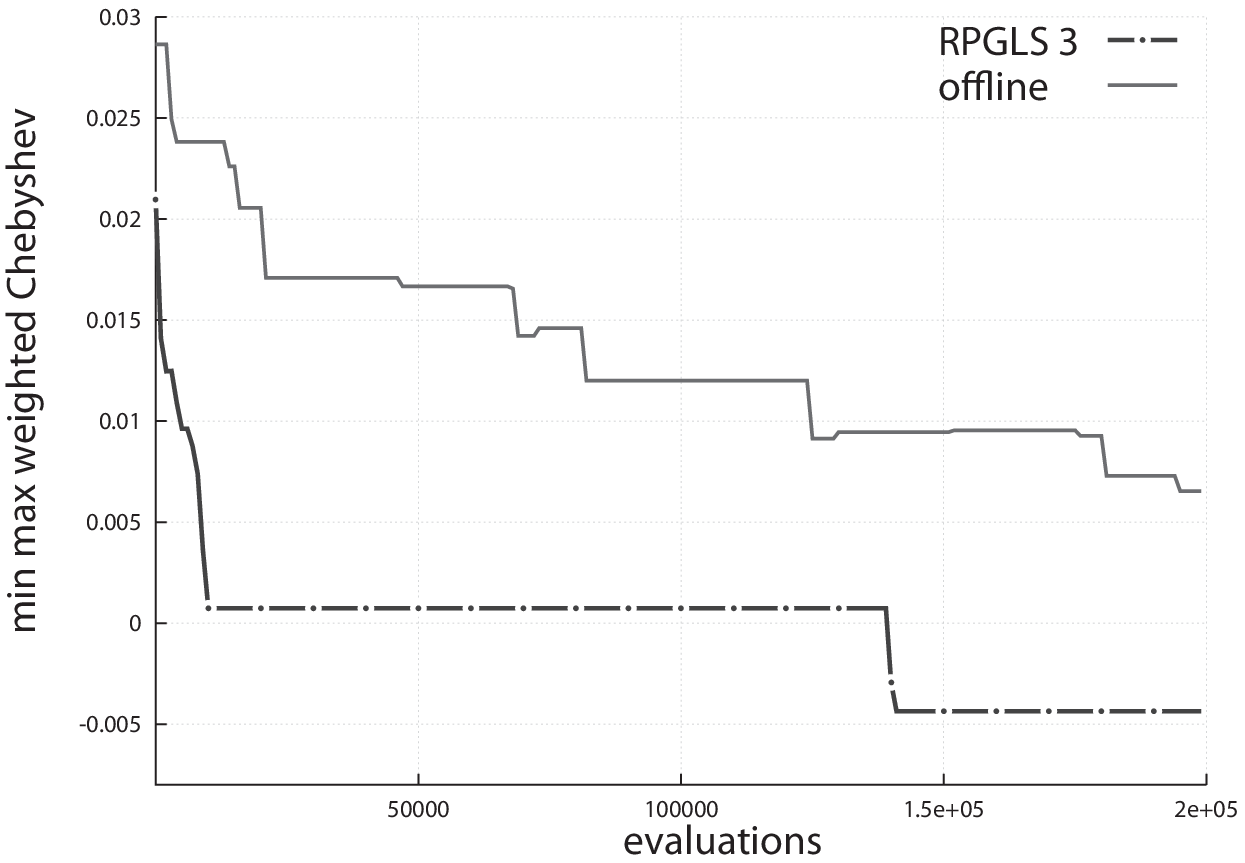}
\end{minipage}
\begin{minipage}[hbt]{8cm}
\centering
  \includegraphics[width=8cm]{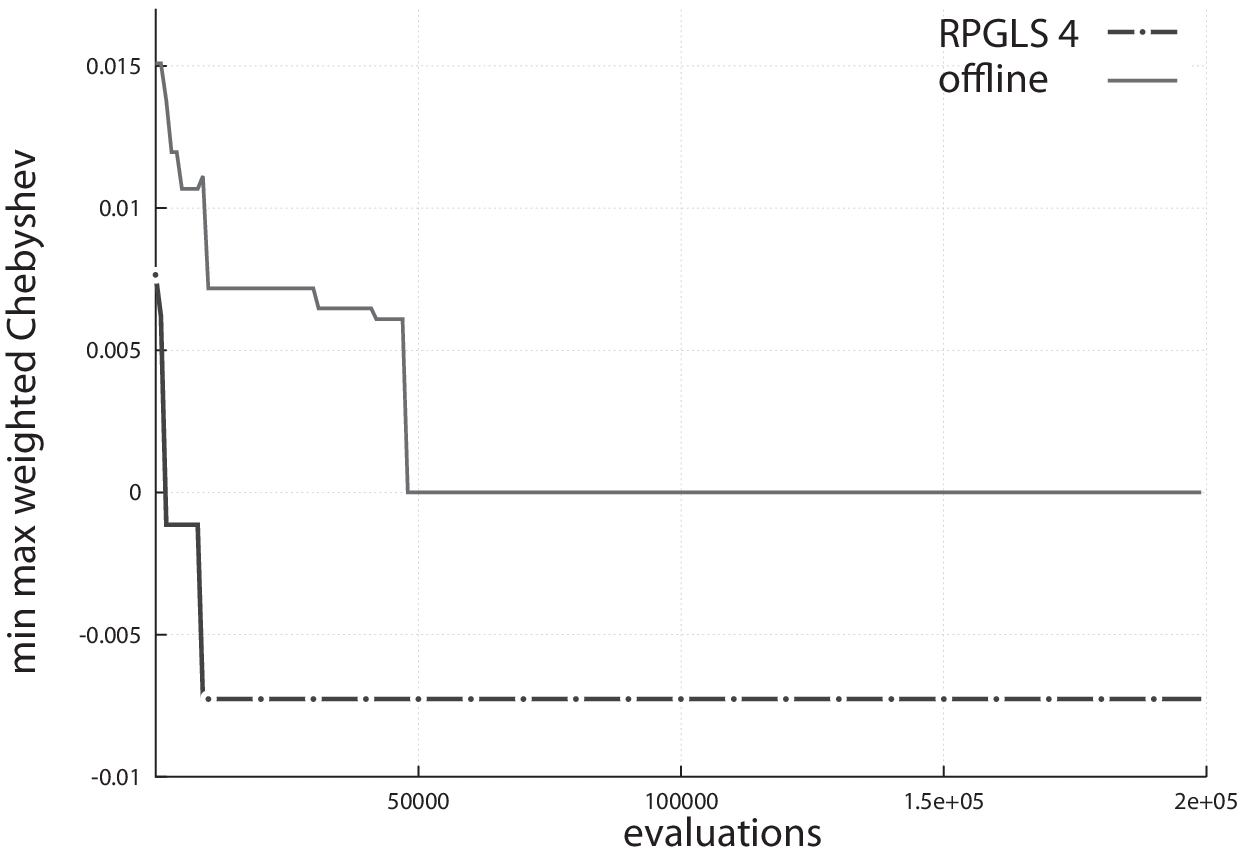}
\end{minipage}

\begin{minipage}[hbt]{8cm}
  \centering
  \includegraphics[width=8cm]{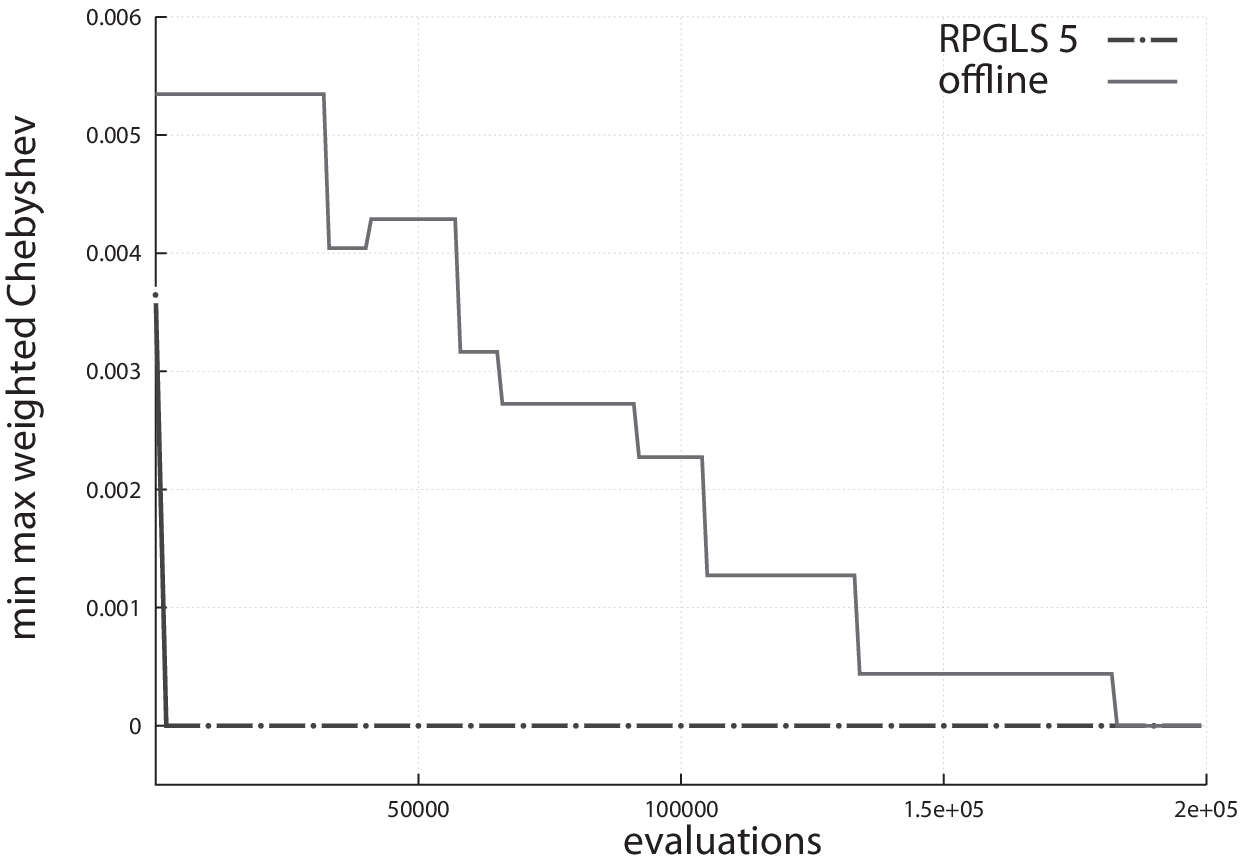}
\end{minipage}
\begin{minipage}[hbt]{8cm}
\centering
  \includegraphics[width=8cm]{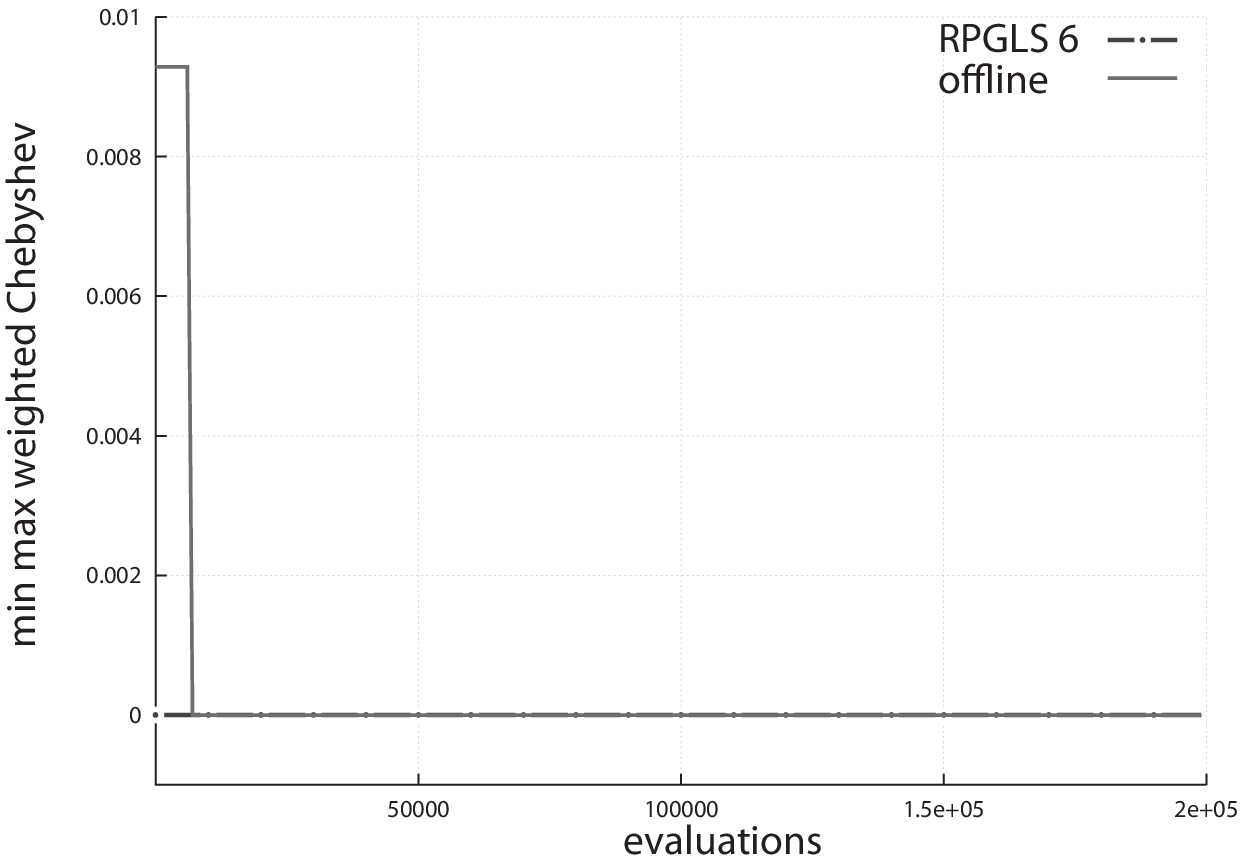}
\end{minipage}
\end{figure}
\begin{figure}
  \includegraphics[width=8cm]{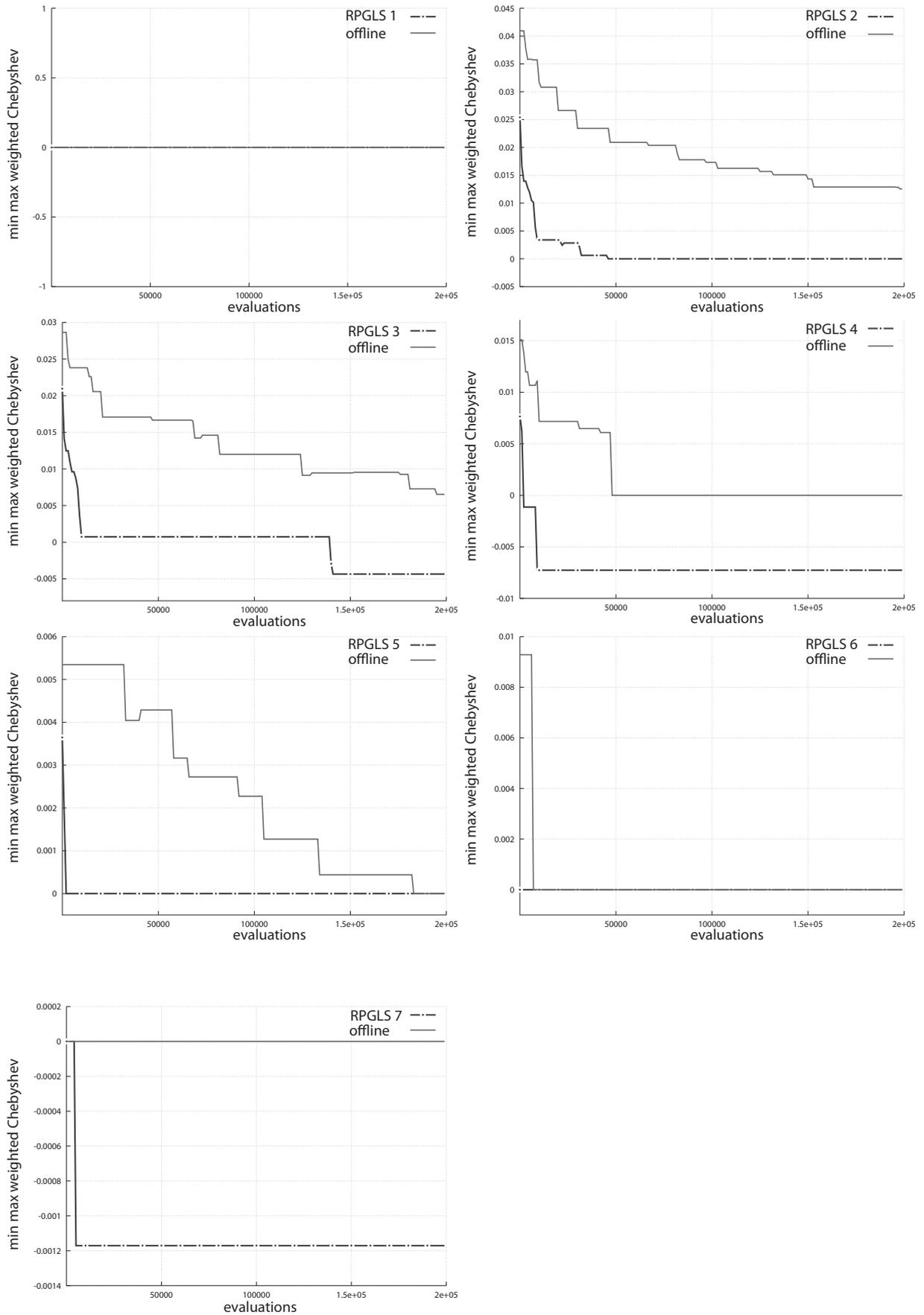}
 \caption{The capabilities of the RPGLS and the offline approach are illustrated for reference points $R_{1}$ to $R_{7}$ ({\tt{GS-01}}). }\label{fig:results_gs-01}
\end{figure}

\begin{table}[]
\centering
\caption{Computational time of test runs for different reference points of {\tt{GS-01}} and {\tt{GS-02}}.}
\label{tbl:computationaltime}
\smallskip
\begin{tabular}{|c|c||c|c|}\hline
\multicolumn{2}{|c|}{{\tt{GS-01}}}& \multicolumn{2}{|c|}{{\tt{GS-02}}}\\\cline{1-4}
 reference point & hours &  reference point & hours \\\hline\hline
 $R_1$& $0.000040$ &  $R_1$ & $0.00013$\\\hline
 $R_2$ & $1.52$   & $R_2$ & $4.99$\\\hline
 $R_3$ &  $1.28$  & $R_3$ &$4.2$\\\hline
 $R_4$ &  $1.23$  & $R_4$ & $3.53$\\ \hline
 $R_5$ & $1.12$ & $R_5$ &$3.36$ \\ \hline
 $R_6$ &  $1.0$ &  $R_6$ & $3.07$\\ \hline
 $R_7$ &  $1.0$ &  $R_7$ & $2.52$\\ \hline
  &  &  $R_8$ & 2,61\\ \hline
\end{tabular}
\end{table}

\begin{figure}[!ht]
\centering
\subfloat[$R_{2}$]
{\includegraphics[width=8cm]{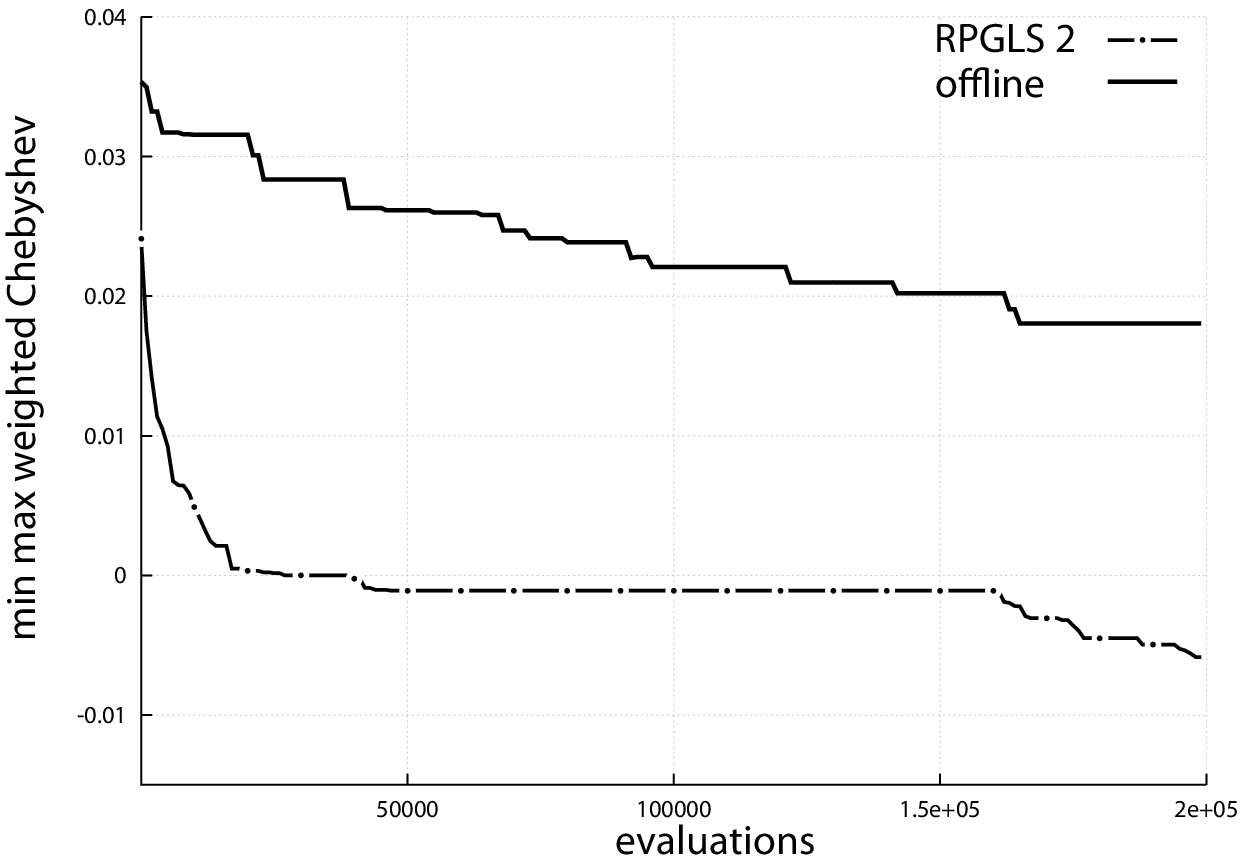}{\label{fig:gs-02-rp2}}}
\subfloat[$R_{8}$]
{  \includegraphics[width=8cm]{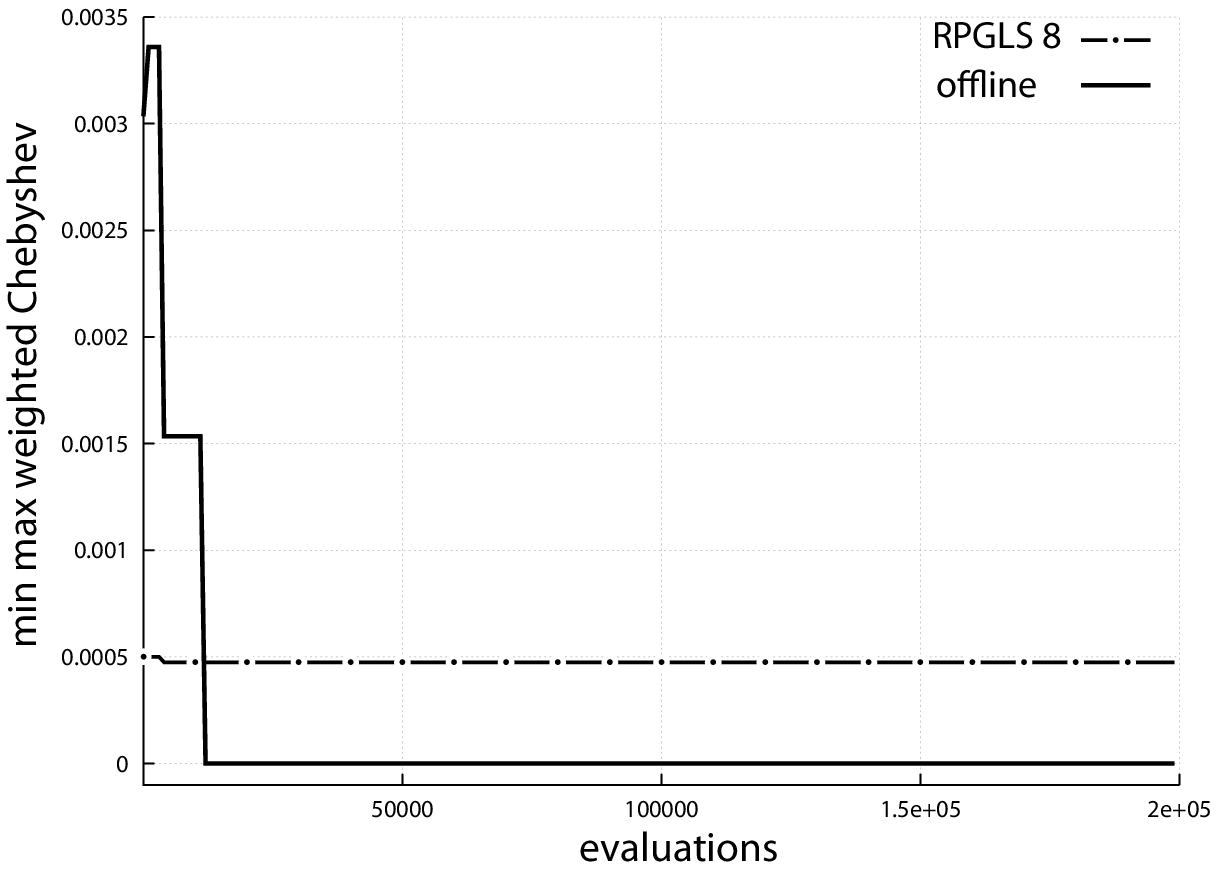}{\label{fig:gs-02-rp8}}}
 \caption{Exemplarily, the capabilities of the RPGLS and the offline approach are illustrated for reference points $R_{2}$ and $R_{8}$ ({\tt{GS-02}}). {\label{fig:results_gs-02}}}
\end{figure}

Figures~\ref{fig:results_gs-01} and~\ref{fig:results_gs-02} show the capabilities of the reference point-based guided local search, and as well of the offline approach. In Figure~\ref{fig:results_gs-01} and~\ref{fig:results_gs-02} both approaches are stopped after 200{,}000 evaluations.

Figure~\ref{fig:results_gs-01} shows the results of the seven reference points of instance {\tt GS-01}. We can see that the relative performance of the RPGLS differs depending on the chosen reference point. Note that we assume that the DM can state his/her preferences and is not indifferent between different reference points. However, the indifference can straightforward applied, by guiding the search into both search directions.

In case of $R_{1}$, $R_{6}$, and $R_{7}$, both approaches, i.\,e.\ the interactive and the offline approach, perform almost identical (with a small difference in case of $R_{7}$). Unfortunately, when $R_{1}$ is selected, both approaches terminate immediately (dashed (RPGLS 1) and solid line are on top of each other in Figure~\ref{fig:results_gs-01}) because they do not find a better solution. For $R_{1}$, this is due to the fact that the `period' vector $\pi = \left( \pi_1, \ldots, \pi_i,\ldots ,\pi_n \right)$ cannot be improved by the savings heuristic and incidentally, a better routing algorithm as reported in~\cite{geiger:2011}, with a more advanced record-to-record travel algorithm~\cite{li:2007:article} must be used to reduce the sum of the routing costs. Also in case of $R_{3}$ and $R_{4}$, only small improvements are possible, and interactive search cannot significantly outperform the offline approach.

Table~\ref{tbl:computationaltime} gives an overview of the computational times for the selected reference point.

Stressing the difference of the approaches, RPGLS could slightly improve the already existing best-known solutions, if $R_{3}$, $R_{4}$ and $R_{7}$ are selected. In the case of $R_{5}$ and $R_{6}$, the RPGLS achieves the same quality as the offline approach, but especially for $R_{2}$ and $R_{5}$ the RPGLS is faster, i.\,e.\ RPGLS finds the most-preferred solution for $R_{5}$ after $2{,}000$ evaluations (computational time of 23.7 seconds) and the offline approach after $183{,}000$ evaluations (57 minutes). It has to be mentioned that the RPGLS is faster for all investigated reference points, except of $R_{1}$ (where no improvement is possible in either approach). As a tendency, the RPGLS achieves a clear speedup of the search, and the investigated outcomes have at least the same quality as the offline approach.

Surprisingly, in some rare cases, we observe that a curve, as e.\,g.\ shown for the solid line of {RPGLS 5} in Figure~\ref{fig:results_gs-01}, can increase by a small value. This appears when the local search slightly improves a solution with respect to the reference point. However, in comparison with the \emph{ex post} best-known solution of the \emph{a posteriori approach}, the weighted distance can slightly worsen. Note that this only occurs because we know the best-known solution of the offline approach in advance (but not known of the investigated algorithm).

The results for the other instance {\tt GS-02} are in line with the ones of {\tt GS-01}. Figure~\ref{fig:results_gs-02} highlights some typical results of {\tt GS-02}. For $R_{2}$, the RPGLS performs clearly better than the offline version, speeding up the solution process in the chosen subregion. In this sense, the observations made for instance {\tt GS-01} are confirmed. In case of $R_{8}$, RPGLS performs worse than the offline approach, but only by a small value. Again, this is in line with the analysis of {\tt GS-01}, where the behavior of the two algorithms appears to be similar when selecting reference points towards the extreme ends in outcome space.

\section{\label{sec:future}Conclusions}
An interactive approach to the IRP has been investigated for several reference points and tested on two benchmark data sets. We have studied the effects of choosing different reference points on the obtained capabilities of RPGLS.

Our applied procedure: first, the DM is given a fast, rough approximation of the Pareto-front. Then, (s)he selects a reference point to guide the local search.

When reviewing the results above, the combination of search and interactive decision making is possible for the investigated IRP. The results of instance {\tt GS-01} and {\tt GS-02} are encouraging in terms of quality and time aspects. Obviously, RPGLS achieves a `speedup' of the search by investigating only preferred subregions. On the one hand, this behavior is particularly present when choosing reference points `in the middle' of the Pareto-front. On the other hand, it is rather difficult to improve solutions at the extreme ends of the Pareto-front. Here, the interactive approach does not give an advantage to the DM.

Despite this progress, open questions arise for future research: although we described promising results, the interactive approach must be tested on larger benchmark instances. This is particularly important as we want to study the performance of the approach in  practical settings. Such settings typically involve more than 75 customers.

It might be helpful for the DM to state several different reference points simultaneously in order to get an impression of outcome space in more detail. This idea can be integrated by guiding the search in more preferred subregions.

We also believe that there is a lack of methodology of how to simulate a decision maker in such an interactive setting. The DM might change his/her preference information during the search and decision making process, e.\,g.\ by stating `new' reference points. How can the changing of reference points be integrated in this approach and which solutions should be saved to support the DM? Also the question appears, what happens when the DM gives inconsistent information.


\bibliography{example}
\bibliographystyle{plain}

\end{document}